\documentclass{article}
\usepackage{spconf,amsmath,graphicx}

\usepackage{times}
\usepackage{soul}
\usepackage{url}
\usepackage[hidelinks]{hyperref}
\usepackage[utf8]{inputenc}
\usepackage[small]{caption}
\usepackage{graphicx}
\usepackage{amsmath}
\usepackage{amssymb}
\usepackage{booktabs}
\usepackage{algorithm}
\usepackage{algorithmic}
\usepackage{color}

\usepackage{kotex} 
\usepackage{booktabs}
\usepackage{mathtools}
\usepackage{caption}
\usepackage{subcaption}
\usepackage{textcomp}
\usepackage{siunitx}

\DeclareMathOperator*{\argmax}{argmax}

\title{Unpriortized Autoencoder for Image Generation}
\name{Jaeyoung Yoo \qquad Hojun Lee \qquad Nojun Kwak$^*$
\thanks{* Corresponding author is Nojun Kwak. This work was supported by ICT Research and Development Program of MSIP/IITP, Korean Government (2017-0-00306). }}
\address{
    Seoul National University \\ 
    \tt\small \{yoojy31, hojun815, nojunk\}@snu.ac.kr
}
\begin{document}
%
\maketitle
\begin{abstract}
In this paper, we treat the image generation task using an autoencoder, a representative latent model. Unlike many studies regularizing the latent variable's distribution by assuming a manually specified prior, we approach the image generation task using an autoencoder by directly estimating the latent distribution. To this end, we introduce `latent density estimator' which captures latent distribution explicitly and propose its structure. Through experiments, we show that our generative model generates images with the improved visual quality compared to previous autoencoder-based generative models.
\end{abstract}
\section{Introduction}
Data generation using an autoencoder is performed by sampling from the latent distribution which is the distribution of the latent variables. Therefore, how to express the latent distribution and how to sample a latent variable is a core problem in the task of extending an autoencoder to a generative model.

\begin{figure}[t]
\begin{center}
\includegraphics[width=0.95\linewidth]{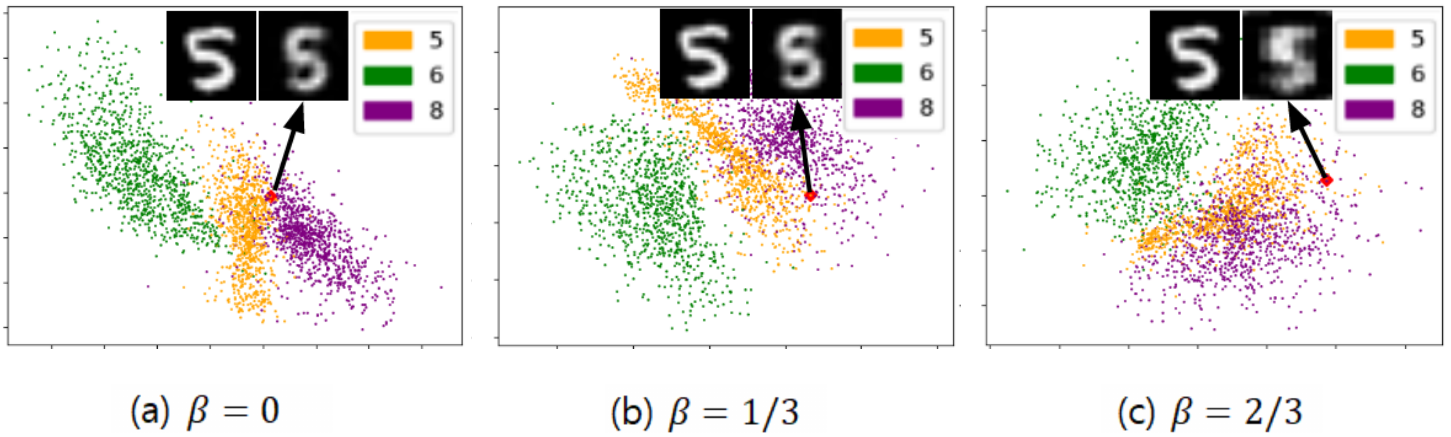}
\end{center}
\vspace{-5mm}
\caption{
    Toy example on MNIST dataset. Two dimensional scatter plot of VAE's latent variables. Each VAE is trained with a different regularization coefficient $\beta$. In each pairs of `5', the left and the right are the target and the reconstructed images, respectively.
}
\vspace{-7mm} 
\label{fig:toy_example}
\end{figure}

Most existing studies, such as \textit{variational autoencoder (VAE)} \cite{kingma2013vae}, approach this problem by assuming a prior as a manually specified distribution (e.g. fully factorized Gaussian). However, manually specified prior may differ from the complex latent nature of the actual data. This difference makes a trade-off between reconstruction and prioritized regularization. In the Figure \ref{fig:toy_example}, we can see that the reconstructed digits are blurry as the weight $\beta$ of the regularization term increases. As a result, it adversely affects the reconstruction and makes a difference between the real latent distribution and the prior, which leads to degradation of the generation quality.

\begin{figure}[t]
\begin{center}
\includegraphics[width=0.8\linewidth]{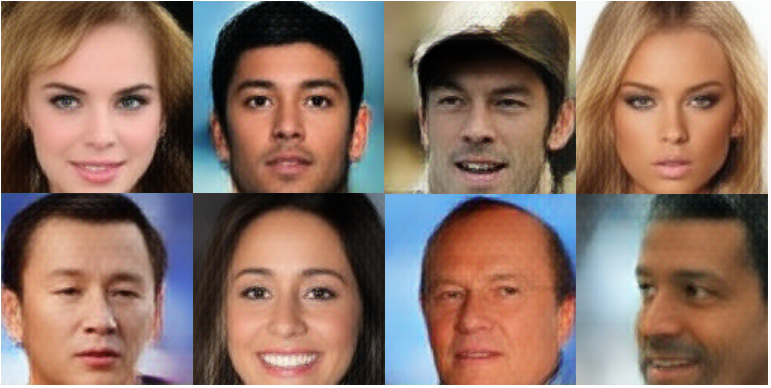}
\end{center}
\vspace{-5mm}
\caption{
    Examples of generated images by our method ($128 \times 128$) using the aligned CelebA dataset.
}
\vspace{-8mm}
\label{fig:example}
\end{figure}

As another approach, we can think of extending autoencoders to generative models by estimating latent distribution from the data. This method is free from the above-mentioned trade-off between the reconstruction and the regularization because the autoencoder can be trained without extra regularization term of the latent distribution. However, in this approach, returning to the nature of the latent variable and the prior, we need to consider a couple of things: (1) The distribution of learned latent variables without prioritized regularization can have a complex form. Therefore, we should be able to model this distribution flexibly. (2) An autoencoder trained solely from data points may be over-fitted or does not learn a proper manifold. The autoencoder should be able to learn meaningful representation via latent variables.

In this paper, we extend the autoencoder to the generative models by estimating the empirical distribution of latent variables obtained from a given dataset using an additional network named as \textit{the Latent Density Estimator (LDE)}. The proposed LDE learns complex distributions flexibly using an auto-regressive approach. Also, in order to preferentially learn the important representation of the given data through the latent variables, we exploit a training strategy that incrementally increases the effective size of the latent vector. As shown in Figure \ref{fig:example}, our method generates clear and diverse images that reflect the characteristics of data well.

\section{Related Works}
A typical study, \textit{variational autoencoder} (VAE) \cite{kingma2013vae}, performs regularization to minimize the KL divergence between the distribution of latent variables and the assumed prior by maximizing the evidence lower bound (ELBO). Another study, \textit{adversarial autoencoder} (AAE) \cite{makhzani2015aae}, uses adversarial training. The encoder is considered as a generator of the latent variables and it is learned to imitate the prior distribution, which acts as a regularization for the latent distribution.

These studies have a drawback in that they must assume a prior in advance of training and must force the latent distribution into a specific form. Normalizing flow \cite{rezende2015normalizingflow} approaches this problem by allowing a more flexible latent distribution to be applied to posterior sequences of invertible transformations. And there are various studies that modified or extended the normalizing flow \cite{kingma2016iaf, papamakarios2017maf,huang2018NAF,dinh2016RealNVP}.

Unlike previous studies, the latent density estimator (LDE) learns the prior by estimating an explicit form of the probability density of the latent variables . In our method, the latent vector is sampled from a latent  distribution $p_{\lambda}(z)$ estimated by LDE.

\section{Latent Density Estimation}
\subsection{Problem formulation}
An autoencoder consists of an encoder and a decoder, which encodes a data point $x$ into a latent vector $z$ ($z = f_{\theta}(x)$) and reconstructs $x$ from $z$ ($x = g_{\psi}(z)$), respectively. Ideally, if a latent vector $z$ can be sampled from a true distribution of $z$, it is possible to generate data using the decoder

Our goal is to directly estimate the latent distribution which represents the dataset well. We deal with this problem by estimating the probability density function of $p(z)$ from dataset using a additional density estimator parameterized by $\lambda$. The density estimator is trained to find $\lambda$ that maximizes the likelihood of $z$ for $\lambda$ as follows.

\begin{equation} \label{eq:max_lambda}
\begin{aligned}
\lambda &= \argmax_{\lambda} \mathbb{E}_{z \sim p_{data}(z)}(\log p_{\lambda}(z)).
\end{aligned}
\end{equation}
Here, the $p_{data}(z)$ is the empirical latent distribution obtained from $p_{data}(x)$ and the encoder $f_\theta{(x)}$ and the LDE estimates $p_{data}(z)$ by $p_{\lambda}(z)$.

\subsection{Overview of the proposed method}
\begin{figure}[t]
\begin{center}
\includegraphics[width=0.88\linewidth]{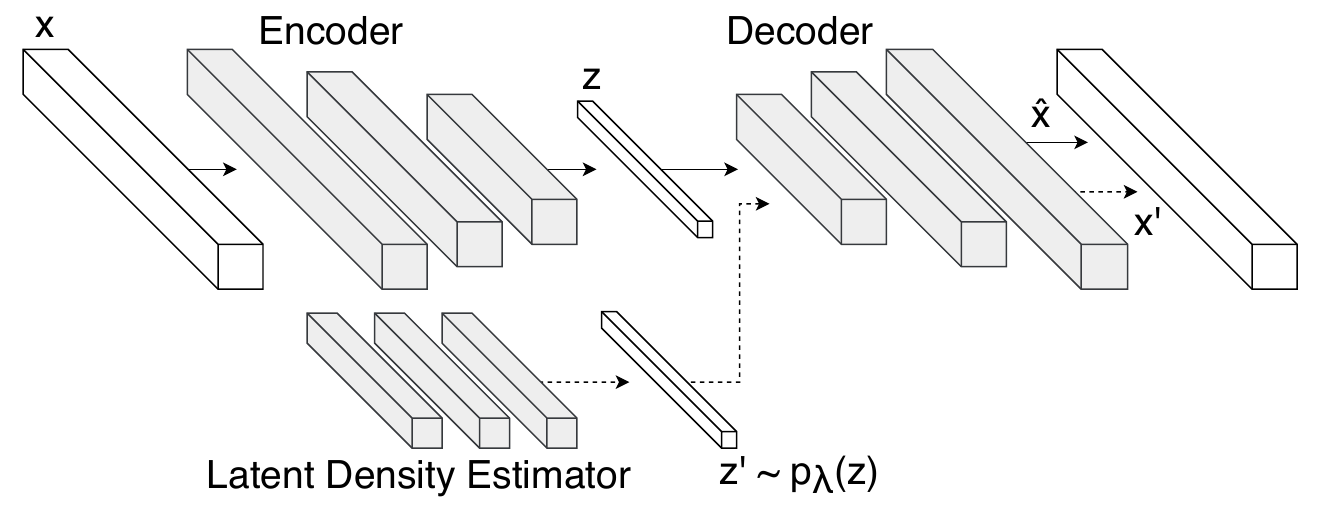}
\end{center}
\vspace{-5mm}
\caption{
    The proposed autoencoder-based generative framework. It is composed of an encoder, a decoder and a Latent Density Estimator (LDE). The solid line represents the inference and the reconstruction path and the dashed line represents the generation path. 
}
\vspace{-5mm}    
\label{fig:architecture}
\end{figure}

Figure \ref{fig:architecture} shows the overall structure of the proposed method for extending autoencoder into generative models. As can be seen in the figure, the architecture of the proposed generative framework consists of an encoder, a decoder and a latent density estimator (LDE).

The encoder and the decoder are used in the form of the deterministic function. 
The LDE expresses the probability density function of $p_{\lambda}(z)$ explicitly. In the generation process, a latent vector $z'$ is randomly sampled from $p_{\lambda}(z)$ and generates new data $x'$ through the decoder.

In this framework, the autoencoder and the LDE are trained according to the following training procedure. First, we train the autoencoder $\hat{x} = g_{\psi}(f_{\theta}(x))$ to reconstruct $x$. Second, after the completion of training the autoencoder, we obtain $p_{data}(z)$ from $p_{data}(x)$ using the encoder $f_{\theta}(x)$. Finally, we train the latent density estimator to estimate $p_{data}(z)$ using $p_{\lambda}(z)$.

\begin{figure}[t]
\begin{center}
\includegraphics[width=0.57\linewidth]{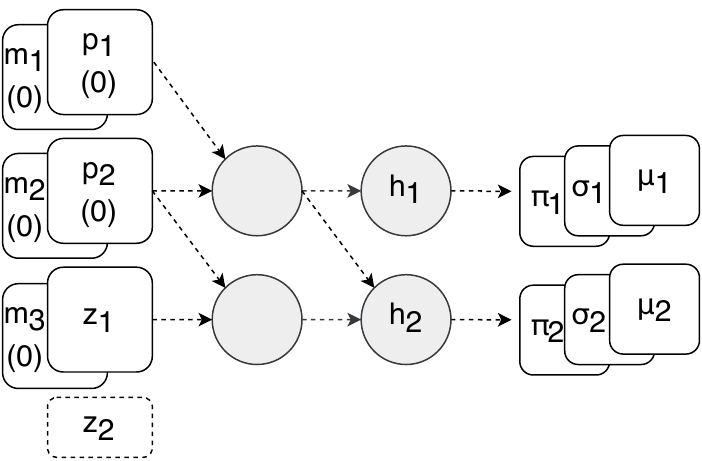}
\end{center}
\vspace{-5mm}
\caption{
    The proposed latent density estimator. Here, the dimension of $z$ is 2, the filter size $s$ is 2. 
}
\vspace{-5mm}
\label{fig:latent_density_estimator}
\end{figure}

\begin{figure*}[t]
\centering
\begin{subfigure}[b]{0.25\textwidth}
    \centering
    \includegraphics[width=\textwidth]{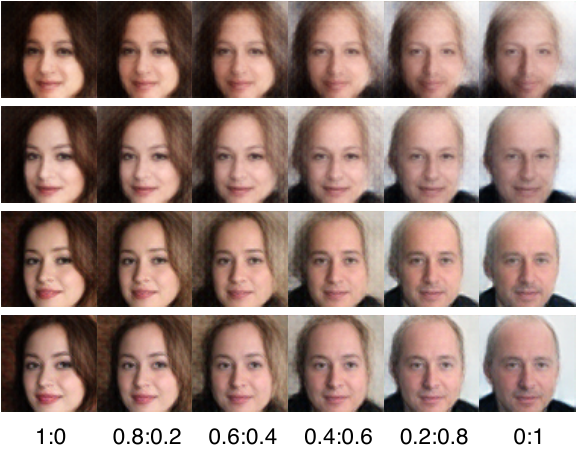}
    \caption{CelebA $\Longleftrightarrow$ CelebA}
\end{subfigure}
\quad
\begin{subfigure}[b]{0.25\textwidth}  
    \centering 
    \includegraphics[width=\textwidth]{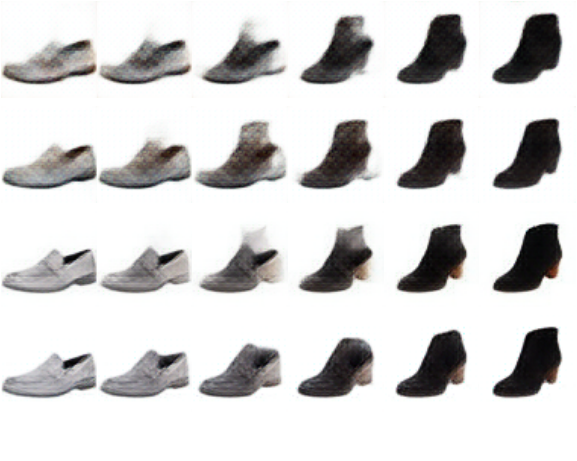}
    \caption{Shoes $\Longleftrightarrow$ Shoes}
\end{subfigure}
\quad
\begin{subfigure}[b]{0.25\textwidth}   
    \centering 
    \includegraphics[width=\textwidth]{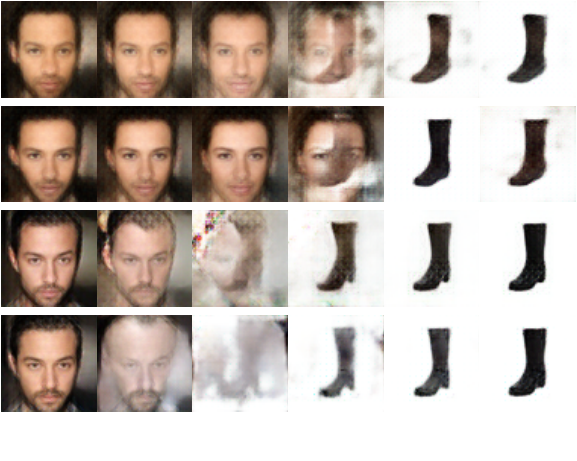}
    \caption{CelebA $\Longleftrightarrow$ Shoes}
\end{subfigure}
\vspace{-4mm}
\caption{
    The interpolation results of three cases in latent space on CelebA and Shoes dastasets. From top to bottom, the first images are result of AAE with Gaussian prior, the second to the fourth are the result of AAE with a mixture of 2 Gaussians, AE, and IAE, respectively.
} 
\label{fig:intepolation}
\vspace{-5mm}
\end{figure*}

\subsection{Latent Density Estimator (LDE)}
\label{section:lde}
The proposed approach of LDE is based on the method in RNADE \cite{uria2013rnade}. If $z$ is a $D$-dimimensional real-valued vector whose $i$-th element is denoted by $z_i$, we factorize $p_{\lambda}(z)$ using the chain rule:

\begin{equation} \label{eq:factorized}
\begin{aligned}
p_{\lambda}(z) &= \prod_{i}^{D} p_{\lambda}(z_{i} | z_{<i}).
\end{aligned}
\end{equation}
Here, $z_{<i}$ is ${z_{1}, ... z_{i-1}}$. The estimated probability density of the $i$-th variable of $z$ is conditionally calculated by the values of the variables with lower indices as $p_{\lambda}(z_{i}|z_{<i})$. It is calculated as follows by the mixture of $K$ univariate Gaussian with its parameters being the mixing coefficient $\pi_{i} = \{\pi_{i,1}, ... \pi_{i,K}\}$, the mean $\mu_{i} = \{\mu_{i,1}, ..., \mu_{i,K}\}$ and the standard deviation $\sigma_{i} = \{\sigma_{i,1}, ..., \sigma_{i,K}\}$:

\begin{equation} \label{eq:mog}
\begin{aligned}
p_{\lambda}(z_{i} | z_{<i}) &= \sum_{k}^{K} \pi_{i,k} \mathcal{N}(z_{i}; \mu_{i,k}, \sigma_{i,k}^{2}). 
\end{aligned}
\end{equation}
Here, the parameters $\pi_{i}$, $\mu_{i}$ and $\sigma_{i}$ are estimated by the LDE. The LDE is learned to find $\pi_{i}$, $\mu_{i}$ and $\sigma_{i}$ by minimizing the loss function which is calculated as the average of all the negative log of (\ref{eq:mog}), $i = 1, \cdots, D$: 

\begin{equation} \label{density_loss}
\begin{aligned}
L_{z} = -\frac{1}{D} \sum_{i}^{D} \log{ p_{\lambda}(z_{i} | z_{<i})}.
\end{aligned}
\end{equation}

As shown in Figure \ref{fig:latent_density_estimator}, the LDE network consists of two parts: dilated causal convolution \cite{oord2016wavenet} and Mixture Density Network \cite{bishop1994mdn}. The dilated causal convolution outputs a series of causal features $h = \{h_{1}, ..., h_{D}\}$ using $z$. Each $h_{i}$ observes $z_{<i}$. The zero vector $p$ is to match the dimension of the input to the first output $h$ to $D$, and $m$ is used to distinguish $z$ from $p$ by setting the masking value as 1 and 0, respectively. The Mixture Density Network estimates the parameters of the $K$ mixture of Gaussians, $\pi_{i}$, $\mu_{i}$ and $\sigma_{i}$, from $h_{i}$ using an $1 \times1$ convolution layer. Here, $\mu_i$ and $\sigma_i$ are obtained by the linear and the exponential activations, respectively, and $\pi_{i}$ is the softmax output for the $K$ Gaussians.

\subsection{Incremental learning of latent vector}
\label{section:iae}
Instead of adding an regularization term to the objective function, we use the structural characteristics of an autoencoder to make an autoencoder learn salient factors of data.
In the training process, initially only a small part of the latent vector is used to learn the autoencoder. Then, as the iteration goes on, the effective size of the latent vector is increased gradually. Here, the unused part of the latent vector is masked to zero. This incremental learning strategy of the latent variables induces the autoencoder to learn the most important representation of data first.

\subsection{Objective Function}
For the training of the autoencoder, we use the distance in the pixel space and the perceptual loss \cite{hou2017percVAE} as follows:

\begin{equation} \label{eq:ae_loss}
\begin{aligned}
L_{z} = L_{MSE}(x, \hat{x}) + \beta L_{Perc}(x, \hat{x}),
\end{aligned}
\end{equation}
where $L_{MSE}(x, \hat{x})$ is the mean sqaured error in the pixel 
space, $L_{Perc}(x, \hat{x})$ is perceptual loss as in 
\cite{hou2017percVAE} and $\beta$ is the weight parameter for the 
perceptual loss.

\section{Experiments}
We will call the autoencoder whose latent vector is trained by applying the incremental learning as IAE. And, AE will denote an autoencoder trained without the incremental learning. For fair comparison, the autoencoders of AAE and ours use the same structure. We use (\ref{eq:ae_loss}) as the reconstruction loss of AAE and ours with the $\beta$ of 0.1. The perceptual loss was calculated using the feature maps at relu11, relu21, and relu31 of Imagenet \cite{imagenet} pretrained VGG19 \cite{simonyan2014vgg}

\begin{figure*}[t]
\centering
\begin{subfigure}[b]{0.25\textwidth}
    \includegraphics[width=\linewidth]{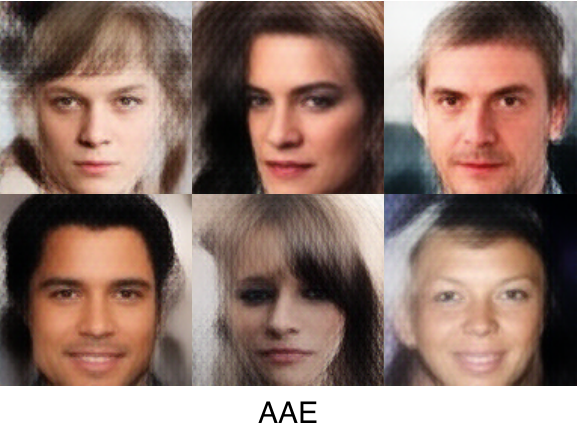}
\end{subfigure}
\quad
\begin{subfigure}[b]{0.25\textwidth}
    \includegraphics[width=\linewidth]{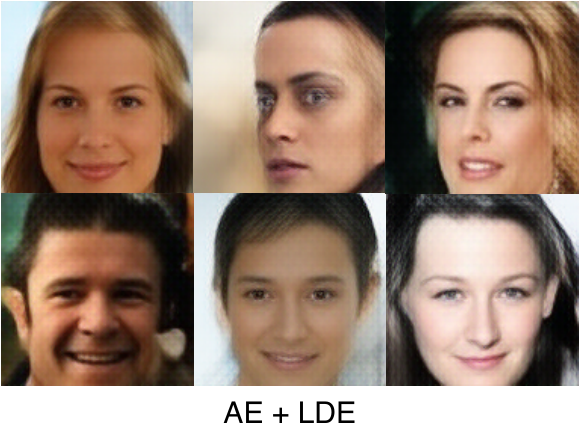}
\end{subfigure}
\quad
\begin{subfigure}[b]{0.25\textwidth}
    \includegraphics[width=\linewidth]{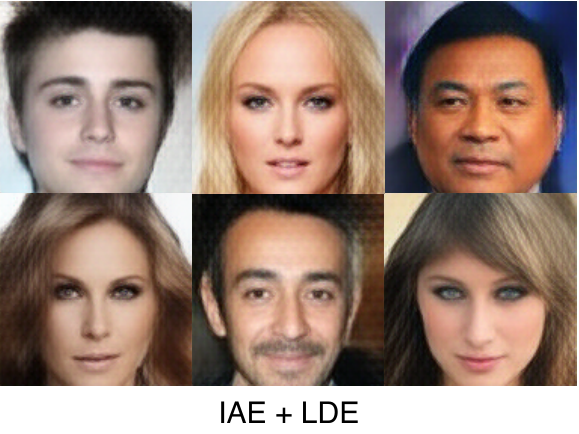}
\end{subfigure}
\vspace{-3mm}
\caption{
    Comparison of random samples from different generative models using autoencoders on CelebA. The generated images from IAE with LDE look the best, most stable and clear.
}
\label{fig:generation}
\vspace{-6mm}
\end{figure*}

\begin{table}[t]
\footnotesize
\def\arraystretch{1.2}
\begin{center}
\begin{tabular}{lc}
    \hline
    Method                                                  & Log Likelihood           \\ 
    \hline \hline
    Deep RNADE (K=10) \cite{uria2014deepRNADE}              & 155.2                     \\
    MADE MoG (K=10) \cite{germain2015made}                  & 153.71$\pm$0.28           \\ 
    Real NVP \cite{dinh2016RealNVP}                         & 153.28$\pm$1.78           \\
    MAF MoG (K=10) \cite{papamakarios2017maf}               & 156.36$\pm$0.28           \\
    MAF-DDSF \cite{huang2018NAF}                            & 157.73$\pm$0.04           \\
    \hline
    LDE (K=1)                                               & 152.68$\pm$0.52           \\
    LDE (K=10)                                              & 157.75$\pm$0.06           \\
    LDE (K=30)                                              & \textbf{158.02$\pm$0.06}  \\
    LDE (K=100)                                             & 157.03$\pm$0.30           \\
    \hline
\end{tabular}
\end{center}
\vspace{-4mm}
\caption{
    Log-likelihood and 2 standard deviations (5 trials) of various density estimation methods on $8 \times 8$ patches of BSDS300 test set. Error bars of 2 standard deviations. 
}
\label{table:likelihood_armdn}
\end{table}

\subsection{Log-likelihood of the Latent Density Estimator}
To quantitatively compare the density estimation performance of the LDE, we measured the log-likelihood in natural images according to the experimental setup of \cite{uria2013rnade,uria2014deepRNADE}. 
We perform the same preprocessing of \cite{uria2013rnade,uria2014deepRNADE} in the BSDS300\cite{MartinFTM01bsds300} dataset. Table \ref{table:likelihood_armdn} shows the log-likelihoods of various density estimators. In this experiment, when $K$ is 10 and 30, the results of our method, LDE, not only outperformed mixture of Gaussian based methods, but also show the best score against all other density estimator compared. In our results, too simple model, LDE with $K$=1, shows the worst performance, and when the model is complex, LDE with $K$=100, the performance becomes lower than that of LDE with $K$=30.

\vspace{-1mm}
\subsection{Latent Space Walking on Two Domains}
In order to understand the representation and structure of the latent space, we perform linearly interpolating two latent vectors. We trained the AAE with a Gaussian prior ($\mathcal{N}(0, I)$), the AAE with a mixture of two Gaussian prior ($\mathcal{N}(-3, I)$ and $\mathcal{N}(3, I)$), AE and IAE that use an 100-dimensional latent vector using the Shoes\cite{shoes1_finegrained} 
and CelebA\cite{liu2015celeba} datasets with a size of $64 \times 64$. Figure \ref{fig:intepolation} shows the results obtained by linearly interpolating two test samples in the latent space for three cases. Figure \ref{fig:intepolation} (a) shows the interpolation on the CelebA and (b) shows the results on the Shoes datasets, both of which are the interpolation in the same domain. Every results of (a) are generally plausible, it shows that the face attributes such as visual age change smoothly. However, in (b), only the results of IAE change the shape continuously. Figure \ref{fig:intepolation} (c) shows the interpolation between different data domains. AAE produces an image in which a person's face and shoes are overlapped during interpolation. In contrast to AAE, our IAE produces images that can not be seen as shoes or human faces in the middle. In fact, the images of a person's face and those of shoes belong to the semantically different data domain from each other, so this gaps can be meaningful for the latent representation of an autoencoder.

\vspace{-1mm}
\subsection{Image Generation}
For qualitative comparison of generation results, We performed image generation on CelebA. The LDE with 30 mixture of Gaussian was applied to AE and IAE. Figure \ref{fig:generation} shows 6 random generation results for AAE, AE with LDE, and IAE with LDE. For a fair comparison, we present successively generated samples, not cherry picking the samples. As with the image quality, the generation results of AAE are generally unclear. AE with LDE produces the more sharp and detailed image compared to AAE, but failure cases are often generated. The results of IAE with LDE show the most stable and best image quality among the compared methods.

\begin{table}[t]
\small
\def\arraystretch{1.1}
\begin{center}
\begin{tabular}{lcc}
\hline
Method                                  & MNIST(10K)            & TFD(10K)              \\ 
\hline \hline
Deep GSN \cite{bengio2014deepGCN}       & 214$\pm$1             & 1890$\pm$29           \\ 
GAN \cite{goodfellow2014GAN}            & 225$\pm$2             & 2057$\pm$26           \\ 
GMMN + AE \cite{li2015gmmn}             & 282$\pm$2             & 2204$\pm$20           \\ 
AAE \cite{makhzani2015aae}              & \textbf{340$\pm$2}    & 2252$\pm$16           \\ 
VAE + IAF \cite{kingma2016iaf}          & 272$\pm$2             & -                     \\         
eVAE \cite{yeung2017eVAE}               & 337$\pm$2             & 2371$\pm$20           \\ 
\hline
AE + LDE                                & 326$\pm$2             & 2476$\pm$33   \\ 
IAE + LDE                               & 326$\pm$2             & \textbf{2507$\pm$32}   \\ 
\hline
\end{tabular}
\caption{
    Test log-likelihood and the standard error of various generative models on MNIST and TFD dataset. Obtained through Parzen window based density estimation.
}
\label{table:likelihood_gen}
\end{center}
\vspace{-10mm}
\end{table}

\subsection{Log-likelihood of Generative Models}
When the log-likelihood can not be computed directly, the Parzen window based density estimation is a commonly used method for evaluation of generative models. We used this evaluation method as specified in \cite{bengio2014deepGCN,li2015gmmn,makhzani2015aae}. 
We trained our generative models without perceptual loss. The dimensionality of the latent vector used was 8 for MNIST\cite{lecun1998mnist} and 15 for TFD\cite{susskind2010tfd}. $K$ of LDE is 30 and the scale parameters of the Gaussian kernel are found through the grid search on the validation set. As can be seen in the table \ref{table:likelihood_gen}, our results show the lower likelihood compared to AAE and eVAE on MNIST, but we achieved the highest log-likelihood far beyond the previous methods on TFDs, relatively a more complex dataset than MNIST.

\section{Conclusion}
In this paper, we tackle the autoencoder-based image generation task without using regularization by a specified prior. We introduced the latent density estimator to estimate the latent distribution. In addition, we proposed an incremental learning strategy of latent vector so that the autoencoder can learn a meaningful representation. The proposed LDE showed better performance than the previous studies in density estimation. As a result, the autoencoder applying the latent density estimator improves the generation quality of the autoencoder based generative models.


\bibliographystyle{IEEEbib}
\bibliography{strings,refs}
\end{document}